\documentclass[11pt,a4paper]{article}

\usepackage[margin=1in]{geometry}
\usepackage{amsmath,amssymb,amsfonts}
\usepackage{booktabs}
\usepackage{graphicx}
\usepackage[hypertexnames=false]{hyperref}
\usepackage{url}
\usepackage{enumitem}
\usepackage{float}
\usepackage[section]{placeins}
\usepackage[protrusion=true,expansion=false]{microtype}

\graphicspath{{figures/}}

\hypersetup{colorlinks=true, linkcolor=blue, citecolor=blue, urlcolor=blue}

\title{%
  \textbf{Eddy-VL 1.9B: Structural Pruning and Layered Distillation\\
  for Edge-Deployable Multimodal Embedding}
}

\author{%
  Hanyeong Cho$^{1}$ \and Changwoo Kim$^{1}$ \and Taeuk Chu$^{1}$ \and Jimin Park$^{1}$ \\[0.4em]
  $^{1}$Urock-AI Lab, Republic of Korea \\
  \url{https://huggingface.co/Urock-AI} \quad \url{https://urock.kr}
}

\date{}

\begin{document}
\maketitle

% ---------------------------------------------------------------------------
\begin{abstract}
In this report, we introduce \textbf{Eddy-VL 1.9B}, a compressed multimodal embedding model
built on \textbf{Qwen3-VL-Embedding-2B}~\cite{qwen3_vl_embedding_2b,li2026qwen3vl_embedding}
for offline, edge-deployable vision--language retrieval.
Eddy-VL targets air-gapped forensic and investigative settings where cloud APIs are
unavailable and latency matters.
Compression combines (i)~probe-driven structural pruning that removes four redundant
text-decoder layers (28$\rightarrow$24) ranked by adjacent-layer linear CKA~\cite{kornblith2019cka},
and (ii)~layered knowledge distillation~\cite{hinton2015distilling} with hole-covering
teacher--student mappings, mid-layer attention-map $1{-}\mathrm{CKA}$, and final-layer
MSE+cosine with Matryoshka dims~\cite{kusupati2022matryoshka} $\{128,\ldots,2048\}$.
The release model contains \textbf{1,926,188,032} parameters (3.85\,GB bf16), a
$\sim$9.5\% reduction from the 2.13B teacher.\footnote{``Eddy-VL 1.9B'' is a release name;
the exact parameter count is 1.93B.}
Empirical evaluations on MMEB-V2~\cite{meng2025mmebv2} (78 tasks, VLM2Vec~\cite{jiang2024vlm2vec})
show that Eddy-VL scores \textbf{63.2} overall vs.\ \textbf{68.9} for the teacher under our
protocol (91.7\% retained), recovering 6.4 of 12.1 points lost by pruning alone (56.8).
Compositional probes remain near teacher quality on SugarCrepe~\cite{hsieh2023sugarcrepe}
(\textbf{86.1} vs.\ \textbf{86.4}), MR\textsuperscript{2}-Bench~\cite{zhou2025mr2bench}
(\textbf{24.5} vs.\ \textbf{24.7}), and ARO~\cite{yuksekgonul2023aro}
(\textbf{59.5} vs.\ \textbf{60.4}); Winoground~\cite{thrush2022winoground} group score is the
main weakness (\textbf{6.8} vs.\ \textbf{8.5}).
Depth pruning also reduces forward latency by $\sim$10\% (150.0$\rightarrow$136.4\,ms/image on
NVIDIA DGX Spark; FlashAttention-2~\cite{dao2022flashattention}).
This report presents the architecture, compression methodology, training procedures, and
evaluation results, demonstrating effectiveness for multimodal retrieval under constrained
edge deployment.
Weights and inference code are released on Hugging Face~\cite{eddy_vl_embedding_19b}.
\end{abstract}

% ===========================================================================
\section{Introduction}
\label{sec:introduction}
% ===========================================================================

Multimodal embedders map images, text, video, and documents into a shared vector space for
nearest-neighbor retrieval.
Recent VLM-based embedders~\cite{jiang2024vlm2vec,li2026qwen3vl_embedding} achieve strong
MMEB-V2~\cite{meng2025mmebv2} scores via contrastive pre-training and reranker distillation on
Qwen3-VL backbones~\cite{bai2025qwen3vl}.
Qwen3-VL-Embedding-2B is a leading open teacher, but its 28-layer text decoder and 4.26\,GB
checkpoint remain heavy for forensic edge settings: air-gapped networks, seized-media processing,
and strict no-cloud policies.

We study \emph{depth compression} of an existing strong teacher---preserving retrieval
behavior, not only parameter count---through probe-driven pruning and layered distillation.
Unlike training a new foundation embedder from scratch, Eddy-VL removes four decoder blocks
and recovers quality with teacher-aligned mid- and final-layer objectives on an explicit
teacher--student layer map that brackets each removed block.

All reported scores use the official MMEB-V2 benchmark via VLM2Vec (78 tasks) and standard
compositional probes.
We report \emph{relative} compression gains under a fixed evaluation protocol rather than
claiming parity with every published leaderboard configuration (\S\ref{sec:eval-setup}).

In the following sections, we describe the model (\S\ref{sec:model}), the training data
(\S\ref{sec:data}), training strategy (\S\ref{sec:training-strategy}), training objectives
(\S\ref{sec:training-objective}), evaluation (\S\ref{sec:eval}), analysis (\S\ref{sec:analysis}),
and conclude with limitations and deployment guidance (\S\ref{sec:conclusion}).

% ===========================================================================
\section{Model}
\label{sec:model}
% ===========================================================================

Eddy-VL and its teacher follow the bi-encoder embedding paradigm introduced in
Qwen3-VL-Embedding~\cite{li2026qwen3vl_embedding}.
Each multimodal instance is encoded independently; relevance is cosine similarity in a shared
2048-dimensional space.
\textbf{Eddy-VL 1.9B} is derived from \textbf{Qwen3-VL-Embedding-2B}~\cite{qwen3_vl_embedding_2b}
by probe-driven structural pruning of the text decoder (28$\rightarrow$24 layers), followed by
layered knowledge distillation from the unchanged teacher checkpoint.

\subsection{Architecture}

Qwen3-VL-Embedding-2B~\cite{qwen3_vl_embedding_2b,li2026qwen3vl_embedding} uses a 24-block
vision encoder ($d{=}1024$, patch 16, merge 2, proj 2048) and a 28-layer text decoder
($d{=}2048$, 16 Q-heads / 8 KV-heads GQA, FFN 6144).
Eddy-VL removes four decoder blocks ($\mathcal{D}=\{3,22,24,25\}$, 0-based HuggingFace
\texttt{layers.*} indices), yielding a 24-layer student with remapped indices.
The vision encoder is frozen during distillation; only the compressed text decoder is trained.
Table~\ref{tab:model-spec} summarizes the release checkpoint.

\begin{table}[!t]
  \centering
  \small
  \caption{Model specifications (teacher vs.\ Eddy-VL 1.9B release).}
  \label{tab:model-spec}
  \begin{tabular}{@{}lcc@{}}
    \toprule
    & Teacher & Eddy-VL 1.9B \\
    \midrule
    Parameters & 2{,}127{,}532{,}032 & 1{,}926{,}188{,}032 \\
    Checkpoint (bf16) & 4.26\,GB & 3.85\,GB \\
    Vision encoder blocks & 24 & 24 (frozen) \\
    Text decoder layers & 28 & 24 \\
    Embedding dimension & 2048 & 2048 \\
    MRL dims & $\{128,\ldots,2048\}$ & $\{128,\ldots,2048\}$ \\
    \bottomrule
  \end{tabular}
\end{table}

\subsection{Embedding method}

The input format follows the Qwen3-VL-Embedding chat template: a system message carries the
retrieval instruction (default: \emph{``Represent the user's input.''}), and the multimodal
instance is passed as a user message (text, image, video, or mixed modalities).
A final \texttt{<|endoftext|>} token is appended; the last hidden state at this token is
L2-normalized to produce the retrieval vector.
The released embedder exposes this pipeline through a self-contained API on Hugging Face.
\begin{figure}[H]
  \centering
  \setlength{\tabcolsep}{3pt}
  \renewcommand{\arraystretch}{1.05}
  \begin{tabular}{|l|}
    \hline
    \scriptsize\ttfamily <|im\_start|>system \\
    \scriptsize\ttfamily [Instruction] \\
    \scriptsize\ttfamily <|im\_end|> \\
    \scriptsize\ttfamily <|im\_start|>user \\
    \scriptsize\ttfamily [Instance] \\
    \scriptsize\ttfamily <|im\_end|> \\
    \scriptsize\ttfamily <|im\_start|>assistant \\
    \scriptsize\ttfamily <|endoftext|> \\
    \hline
  \end{tabular}
  \vspace{-0.4em}
  \caption{Input template for embedding. A system message carries the retrieval instruction,
           and the user message contains the multimodal instance (text/image/video). We pool
           the last hidden state at \texttt{<|endoftext|>} and L2-normalize it.}
  \label{fig:input-template}
\end{figure}
\vspace{-0.4em}

Unless otherwise stated, we use the full 2048-dimensional embedding for retrieval and evaluation
(MRL enables computing lower-dimensional prefixes with the same encoder when needed).

% ===========================================================================
\section{Data}
\label{sec:data}
% ===========================================================================

Distillation uses two proprietary corpora with complementary supervision
(Table~\ref{tab:dataset-modes}):
\textbf{Korean COCO} supplies image--caption pairs---each sample includes a Korean text
caption alongside the image (\S\ref{sec:kococo-corpus});
\textbf{OurDataset} is image-only---roughly 45k RGB images with \emph{no} accompanying text
(\S\ref{sec:our-dataset}).
Both streams share the default retrieval instruction and last-token pooling; only Korean COCO
instances pass caption tokens through the text decoder.
Ablation and release checkpoints train on the combined mixture.

\begin{table}[H]
  \centering
  \small
  \caption{Distillation corpora (approximate instance counts). Korean COCO = AI Hub Korean
           Image Captioning Dataset~\cite{aihub_kococo} (image + Korean caption); OurDataset =
           image only (no text).}
  \label{tab:dataset-modes}
  \begin{tabular}{@{}llp{4.8cm}lr@{}}
    \toprule
    Corpus & Supervision & Sample format & Text? & Approx.\ $N$ \\
    \midrule
    Korean COCO & Caption & image + Korean caption & yes & ${\sim}$410k \\
    OurDataset & None & image only & no & ${\sim}$45k \\
    Combined & Mixed & both formats above & --- & ${\sim}$455k \\
    \bottomrule
  \end{tabular}
\end{table}

\subsection{Korean COCO captions}
\label{sec:kococo-corpus}

\textbf{Korean COCO} denotes the \emph{Korean Image Captioning Dataset} released on
\textbf{AI Hub}~\cite{aihub_kococo,aihub_korea} (dataset no.\ 261): MS COCO~\cite{lin2014coco}
images paired with \textbf{Korean text captions} obtained by machine-translating the original
English COCO captions, followed by AI Hub quality correction.
Each record provides one or more \texttt{caption\_ko} strings per image (JSON metadata with
UTF-8 text, as distributed by AI Hub).
We use a processed training split with on the order of \textbf{80k} unique images; with
multiple captions per image this expands to roughly \textbf{410k} distillation instances.
Every instance is a multimodal forward pass: image tokens plus a Korean caption under the
default retrieval instruction.
The corpus is licensed through AI Hub and is \textbf{not} redistributed with Eddy-VL.

\subsection{OurDataset}
\label{sec:our-dataset}

OurDataset is a proprietary \textbf{synthesized image corpus} for forensic and investigative
retrieval, built by merging curated subsets from multiple sources:
licensed \textbf{AI Hub} resources~\cite{aihub_korea} (office-document generation, Korean
visual understanding, multimodal information retrieval, and academic/paper understanding),
public benchmarks (MS COCO~\cite{lin2014coco}, Korean COCO~\cite{aihub_kococo},
SUN~\cite{xiao2010sun},
RVL-CDIP~\cite{harley2015rvlcdip}, CORD v2~\cite{park2019cord}), and internally collected or
synthetically generated media.
Unlike Korean COCO, \textbf{no text captions are provided}: each sample is a single image
with the default retrieval instruction only (image-only distillation).

The corpus spans forensic-relevant domains including scenes and places, people and animals,
email, academic papers, presentations and slides, handwriting, screenshots, administrative
and financial documents, road imagery, device and PC/phone screens, identity documents,
technical drawings, logistics, receipts, GPS/geolocation imagery, and whiteboard captures.
Overall volume is roughly \textbf{45k} images.
Neither OurDataset nor AI Hub--derived subsets (including Korean COCO~\cite{aihub_kococo})
are publicly redistributed.
Source-level attributions (AI Hub, Korean COCO, MS COCO, SUN, RVL-CDIP, CORD) appear in the
prose above;
Table~\ref{tab:dataset-modes} reports only corpus scale and supervision type.

% ===========================================================================
\section{Training strategy}
\label{sec:training-strategy}
% ===========================================================================

\begin{figure}[H]
  \centering
  \includegraphics[width=\textwidth]{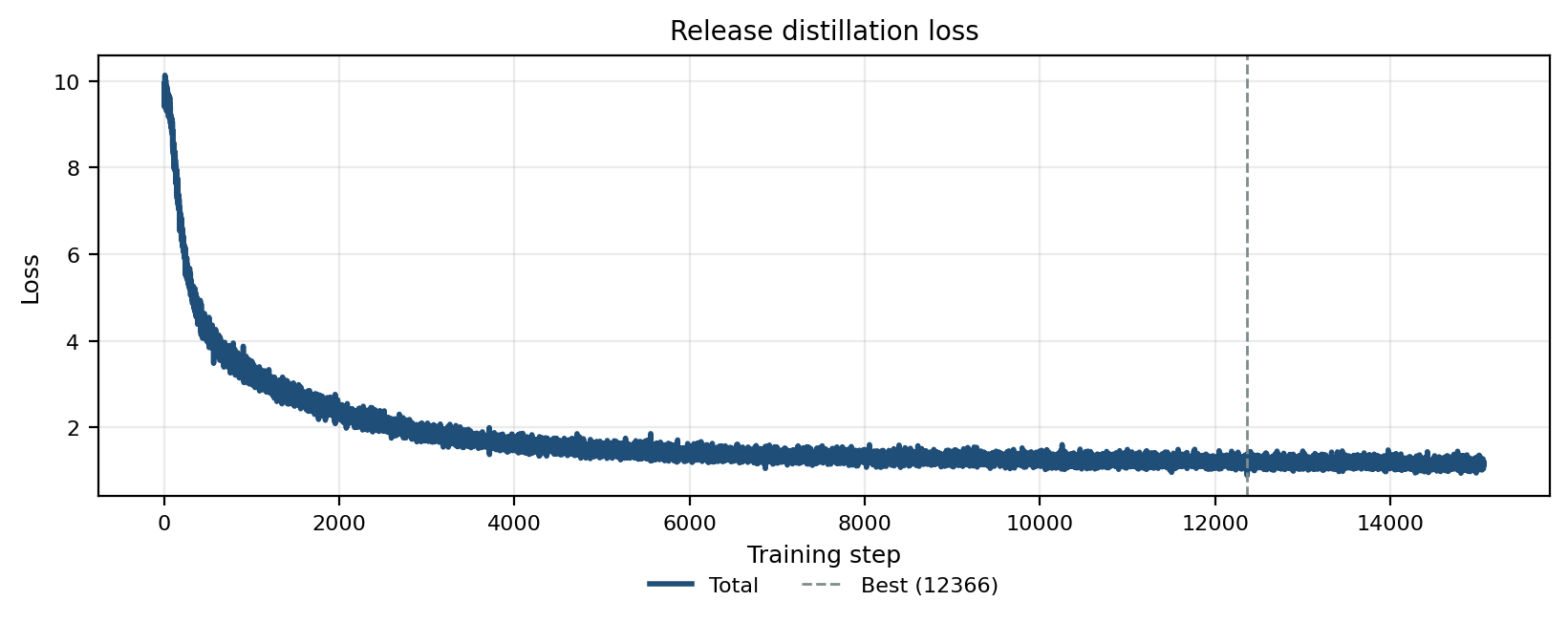}
  \caption{Release distillation loss curve (best checkpoint highlighted).}
  \label{fig:loss}
\end{figure}

Training proceeds in two stages: probe-driven structural pruning (\S\ref{sec:pruning}),
then distillation from the unchanged teacher checkpoint on Korean COCO + OurDataset.
Figure~\ref{fig:loss} shows the training loss curve; the best checkpoint is used for all
reported Eddy-VL results.

\subsection{Probe-driven structural pruning}
\label{sec:pruning}

\begin{figure}[H]
  \centering
  \includegraphics[width=\textwidth]{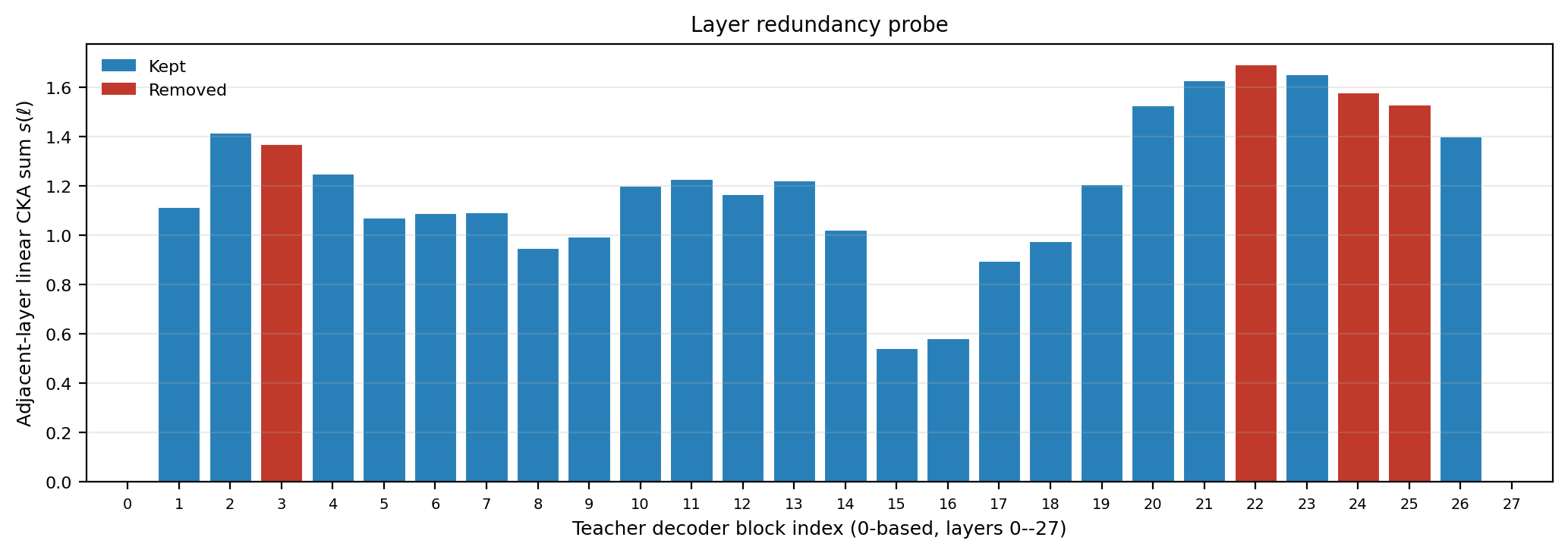}
  \caption{Teacher layer probe: $s(\ell)$ vs.\ 0-based decoder block index
           $\ell\in\{0,\ldots,27\}$.
           Red bars: removed blocks $\mathcal{D}=\{3,22,24,25\}$.}
  \label{fig:probe}
\end{figure}

On $N$ calibration image--caption pairs, extract last-token hiddens after each of the 28
teacher decoder blocks.
We use \textbf{0-based block indices} $\ell\in\{0,\ldots,27\}$ (HuggingFace
\texttt{language\_model.layers[$\ell$]}), form $X_\ell\in\mathbb{R}^{N\times H}$, and score
adjacent-layer redundancy with linear CKA~\cite{kornblith2019cka}:
\(s(\ell)=\mathrm{CKA}(X_{\ell-1},X_\ell)+\mathrm{CKA}(X_\ell,X_{\ell+1})\)
(middle blocks only).
We rank blocks on $N{=}1{,}000$ calibration pairs (Korean COCO + OurDataset mix).
Greedy top-4 CKA would remove $\{21,22,23,24\}$; we use $\mathcal{D}=\{3,22,24,25\}$
instead (block 3 replaces block 23 for early-depth coverage).
Figure~\ref{fig:probe} plots $s(\ell)$ across all decoder blocks.

\paragraph{Beyond the $s(\ell)$ sum.}
Ranking only by $s(\ell)$ also suggests an alternative four-block set
$\{2,21,22,24\}$ (blocks 2, 21, 22, and 24 rank 7th, 3rd, 1st, and 4th among middle
blocks).
We therefore inspect the \emph{decomposition}
$s(\ell)=\mathrm{CKA}(X_{\ell-1},X_\ell)+\mathrm{CKA}(X_\ell,X_{\ell+1})$ rather than
the sum alone (Table~\ref{tab:pruning-balance}).
% Auto-generated CKA balance table
\begin{table}[H]
\centering
\small
\caption{Adjacent CKA decomposition for candidate blocks ($N{=}1{,}000$ calibration pairs). Balance $= 1 - |\mathrm{CKA}_{\ell-1,\ell} - \mathrm{CKA}_{\ell,\ell+1}|/s(\ell)$; higher is more symmetric.}
\label{tab:pruning-balance}
\begin{tabular}{@{}rcccc@{}}
\toprule
Block & $\mathrm{CKA}_{\ell-1,\ell}$ & $\mathrm{CKA}_{\ell,\ell+1}$ & $s(\ell)$ & Balance \\
\midrule
2 & 0.746 & 0.668 & 1.414 & 0.944 \\
3$^\ast$ & 0.668 & 0.702 & 1.370 & 0.975 \\
21 & 0.803 & 0.825 & 1.629 & 0.986 \\
22$^\ast$ & 0.825 & 0.865 & 1.691 & 0.976 \\
23 & 0.865 & 0.786 & 1.652 & 0.952 \\
24$^\ast$ & 0.786 & 0.792 & 1.579 & 0.996 \\
25$^\ast$ & 0.792 & 0.735 & 1.528 & 0.963 \\
\bottomrule
\end{tabular}
\\[0.25em]
\footnotesize $^\ast$Removed in $\mathcal{D}=\{3,22,24,25\}$.
\end{table}

Block~2 has a high $s(\ell)$ because \emph{both} adjacent CKAs are elevated
($0.746$ below, $0.668$ above), not because one side dominates; its balance score
($1 - |\mathrm{CKA}_{\ell-1,\ell}-\mathrm{CKA}_{\ell,\ell+1}|/s(\ell)=0.944$) is only
moderately asymmetric.
However, block~2 is \emph{more} similar to block~1 than to block~3
($\mathrm{CKA}_{1,2}>\mathrm{CKA}_{2,3}$), whereas block~3 is more balanced
(balance $0.975$) and better preserves the early-depth transition.
The neighbor block~1 shows the largest local imbalance in the stack
($0.368$ vs.\ $0.746$), indicating a sharp early representation ramp that makes
shallow-layer removal risky even when $s(\ell)$ is high.

We validate the layer choice with a structural-pruning ablation (weight remapping only,
no distillation): export a 24-layer student with $\{2,21,22,24\}$ removed and score it
on MMEB-Image (batch size 128; Table~\ref{tab:pruning-layer-ablation}).
% Auto-generated structural pruning MMEB-Image ablation
\begin{table}[H]
\centering
\small
\caption{Preliminary MMEB-Image structural-pruning ablation (8 tasks completed; no distillation, batch size 128). Alternative set $\{2,21,22,24\}$ vs.\ release pruned $\{3,22,24,25\}$.}
\label{tab:pruning-layer-ablation}
\begin{tabular}{@{}lrrr@{}}
\toprule
Task & $\{2,21,22,24\}$ & $\{3,22,24,25\}$ & $\Delta$ \\
\midrule
HatefulMemes & 49.2 & 59.2 & -10.0 \\
ImageNet-1K & 0.1 & 44.3 & -44.2 \\
ImageNet-A & 1.6 & 30.3 & -28.7 \\
ImageNet-R & 0.9 & 69.7 & -68.8 \\
N24News & 4.5 & 25.3 & -20.8 \\
Place365 & 0.3 & 28.1 & -27.8 \\
SUN397 & 0.3 & 52.1 & -51.8 \\
VOC2007 & 3.0 & 80.0 & -77.0 \\
\midrule
Macro-avg (8 tasks) & 7.5 & 48.6 & -41.1 \\
\bottomrule
\end{tabular}
\end{table}

The alternative set collapses on classification-heavy tasks (e.g., ImageNet-1K
$0.1$ vs.\ $44.3$ hit@1), confirming that probe sum alone is insufficient and
supporting $\mathcal{D}=\{3,22,24,25\}$ for the release pruned student.

\subsection{Pipeline summary}

\begin{enumerate}[leftmargin=*]
  \item Rank teacher layers by $s(\ell)$ on $N{=}1{,}000$ calibration pairs.
  \item Export a 24-layer pruned student by weight remapping (no fine-tuning).
  \item Distill the pruned student from the frozen-vision teacher on Korean COCO + OurDataset.
  \item Select the best checkpoint and release it as Eddy-VL 1.9B.
\end{enumerate}

\begin{table}[H]
  \centering
  \small
  \caption{Release distillation hyperparameters.}
  \label{tab:hyperparams}
  \begin{tabular}{@{}ll@{}}
    \toprule
    Setting & Value \\
    \midrule
    Distillation & Layered (attention-map mid, MRL final) \\
    Mid / final loss & $1{-}\mathrm{CKA}$ / MSE+cosine (0.3/0.7) \\
    MRL dims & 128, 256, 512, 1024, 2048 \\
    Layer mapping & $3{:}4,21{:}23,22{:}26,23{:}27,24{:}28$ \\
    Batch / LR / warmup & 32 / $3.63{\times}10^{-5}$ / 0.05 \\
    Max length / image pixels & 8192 / 1024--921600 \\
    Vision encoder & Frozen (no LoRA) \\
    Steps / best checkpoint & 15068 / 12366 \\
    Best total loss & 0.889 \\
    \bottomrule
  \end{tabular}
\end{table}

% ===========================================================================
\section{Training objective}
\label{sec:training-objective}
% ===========================================================================

All distillation runs start from the same 24-layer pruned student with a frozen vision
encoder and a hole-covering teacher--student mapping
$\mathcal{M}=\{(3,4),(21,23),(22,26),(23,27),(24,28)\}$ (student indices 0-based; teacher
indices use the probe hidden-state numbering, teacher block $b$ $\leftrightarrow$ index $b{+}1$)
that pairs each removed teacher block with neighboring survivor layers.
We compare two distillation families: \textbf{hidden} distillation and \textbf{layered}
distillation.

\paragraph{Hidden distillation (final hidden only).}
Hidden distillation aligns \textbf{only the final hidden states}: teacher and student
last-token embeddings are matched with MSE+cosine. No losses are applied on intermediate
layers in $\mathcal{M}$.

\paragraph{Layered distillation (mid + final).}
Layered distillation adds \textbf{mid-layer} losses on each mapped pair in $\mathcal{M}$,
aligning intermediate teacher and student activations (default: $1{-}\mathrm{CKA}$~\cite{kornblith2019cka}
on flattened attention maps), plus a \textbf{final-layer} loss on $(24,28)$ using
MSE+cosine with Matryoshka Representation Learning~\cite{kusupati2022matryoshka}.
Dynamic loss scaling targets a 40/60 mid:final ratio.
The release model uses attention-map CKA mid-losses.
Alternative mid-layer objectives include RCKA~\cite{zhou2024rcka}, STRUCTURE~\cite{groger2025structure},
and mutual-kNN~\cite{park2019rkd}.

\paragraph{Mid-layer objective variants.}
All variants below replace only the \emph{mid} term under layered distillation (the final-layer
MRL MSE+cosine term is kept the same).
\textbf{RCKA}~\cite{zhou2024rcka} matches relation-centered similarity by comparing centered Gram
matrices of teacher vs.\ student representations.
\textbf{STRUCTURE}~\cite{groger2025structure} aligns neighborhood structure by applying a
row-softmax to Gram similarities and minimizing a Jensen--Shannon divergence over the resulting
transition distributions.
\textbf{Mutual-kNN}~\cite{park2019rkd} builds a mutual $k$-nearest-neighbor graph on teacher
features within the batch and minimizes MSE over student vs.\ teacher pairwise cosine similarities
on the mutual edges.

% ===========================================================================
\section{Evaluation}
\label{sec:eval}
% ===========================================================================

\subsection{Setup}
\label{sec:eval-setup}

We evaluate on the official MMEB-V2 benchmark~\cite{meng2025mmebv2} with the public VLM2Vec
harness~\cite{jiang2024vlm2vec} (78 tasks), following the
framework of Qwen3-VL-Embedding~\cite{li2026qwen3vl_embedding}.
MMEB-V2 overall is the unweighted mean of 78 per-task metrics (36 image + 18 video Hit@1;
24 VisDoc nDCG@5).
Fine-grained probes (SugarCrepe, ARO, Winoground, MR\textsuperscript{2}) are a \emph{separate}
compositional suite with cosine retrieval accuracy or nDCG@10;
they do not overlap with MMEB-V2 task definitions.

\paragraph{Hardware.}
Unless noted otherwise, all benchmark scores, fine-grained probes, and forward-pass latency
measurements in this report were run on a single \textbf{NVIDIA DGX Spark} GPU (bf16,
FlashAttention-2 where applicable).

\paragraph{Protocol caveats.}
Our VLM2Vec evaluation uses a fixed generic instruction
(\emph{``Represent the user's input.''}) and 8 video frames per clip.
Qwen's published MMEB numbers use task-tuned prompts and up to 64 frames.
Our teacher re-evaluation scores \textbf{68.9} overall vs.\ \textbf{73.2} on the Qwen model
card~\cite{qwen3_vl_embedding_2b,li2026qwen3vl_embedding}.
We therefore report \emph{relative} gains (68.9$\rightarrow$56.8$\rightarrow$63.2), not
parity with the public leaderboard.

\subsection{MMEB-V2 results}
\label{sec:mmeb-results}

Pruning costs 12.1 points; layered CKA distillation recovers 6.4
(68.9 $\rightarrow$ 56.8 $\rightarrow$ 63.2).
Table~\ref{tab:mmeb-overall} lists headline scores; the 73.2 teacher figure is the
public Qwen3-VL-Embedding-2B leaderboard entry~\cite{qwen3_vl_embedding_2b,li2026qwen3vl_embedding}.

\begin{table}[t]
  \centering
  \caption{MMEB-V2 headline results.}
  \label{tab:mmeb-overall}
  \begin{tabular}{@{}lcccc@{}}
    \toprule
    Model & Overall & Image & Video & VisDoc \\
    \midrule
    Teacher (public leaderboard) & 73.2 & --- & --- & --- \\
    Teacher (our eval.) & 68.9 & 71.0 & 55.2 & 76.1 \\
    Pruned baseline & 56.8 & 55.7 & 46.2 & 66.4 \\
    \textbf{Eddy-VL 1.9B} & \textbf{63.2} & \textbf{63.4} & \textbf{50.0} & \textbf{72.7} \\
    \bottomrule
  \end{tabular}
\end{table}

Eddy retains 91.7\% of teacher MMEB under our protocol while removing 4 decoder layers.
Eddy recovers a substantial portion of the prune-only gap, with especially strong retention
on fine-grained compositional probes (\S\ref{sec:analysis}).

\subsection{Fine-grained probes}

\begin{table}[t]
  \centering
  \small
  \caption{Fine-grained benchmarks (headline metrics).}
  \label{tab:finegrained}
  \begin{tabular}{@{}lcccc@{}}
    \toprule
    Benchmark & Metric & Teacher & Pruned baseline & Eddy \\
    \midrule
    SugarCrepe & Overall (\%) & 86.4 & 75.7 & 86.1 \\
    ARO & Overall (\%) & 60.4 & 61.7 & 59.5 \\
    Winoground & Group (\%) & 8.5 & 5.0 & 6.8 \\
    MR$^2$ & nDCG@10 & 24.7 & 20.3 & 24.5 \\
    \bottomrule
  \end{tabular}
\end{table}

Eddy is within 0.3 points of teacher on SugarCrepe and within 0.2 on MR$^2$ (Table~\ref{tab:finegrained}),
indicating near-full recovery on key compositional probes.

\subsection{Efficiency}
\label{sec:efficiency}

Forward latency (Table~\ref{tab:latency}) was measured on NVIDIA DGX Spark with
FlashAttention-2~\cite{dao2022flashattention} on a held-out internal image set
(\(\sim\)44k images; batch size 24; single GPU).
Eddy-VL matches the pruned baseline architecture; depth pruning yields $\sim$10\% lower
forward time vs.\ the teacher.

% Auto-generated forward latency table
\begin{table}[t]
\centering
\small
\caption{Forward-pass latency on NVIDIA DGX Spark (held-out internal image set, \(\sim\)44k images; FlashAttention-2; batch size 24; single GPU). Depth pruning yields $\sim$10\% lower forward time.}
\label{tab:latency}
\begin{tabular}{@{}lrr@{}}
\toprule
Model & Decoder layers & Forward mean (ms/image) \\
\midrule
Qwen3-VL-Embedding-2B (teacher) & 28 & 150.0 \\
Pruned baseline & 24 & 136.4 \\
\midrule
Speedup (pruned vs.\ teacher) & & 1.10$\times$ ($\sim$10\% faster) \\
\bottomrule
\end{tabular}
\end{table}

% ===========================================================================
\section{Analysis}
\label{sec:analysis}
% ===========================================================================

\subsection{Compression and distillation ablations}

\begin{figure}[H]
  \centering
  \includegraphics[width=\textwidth]{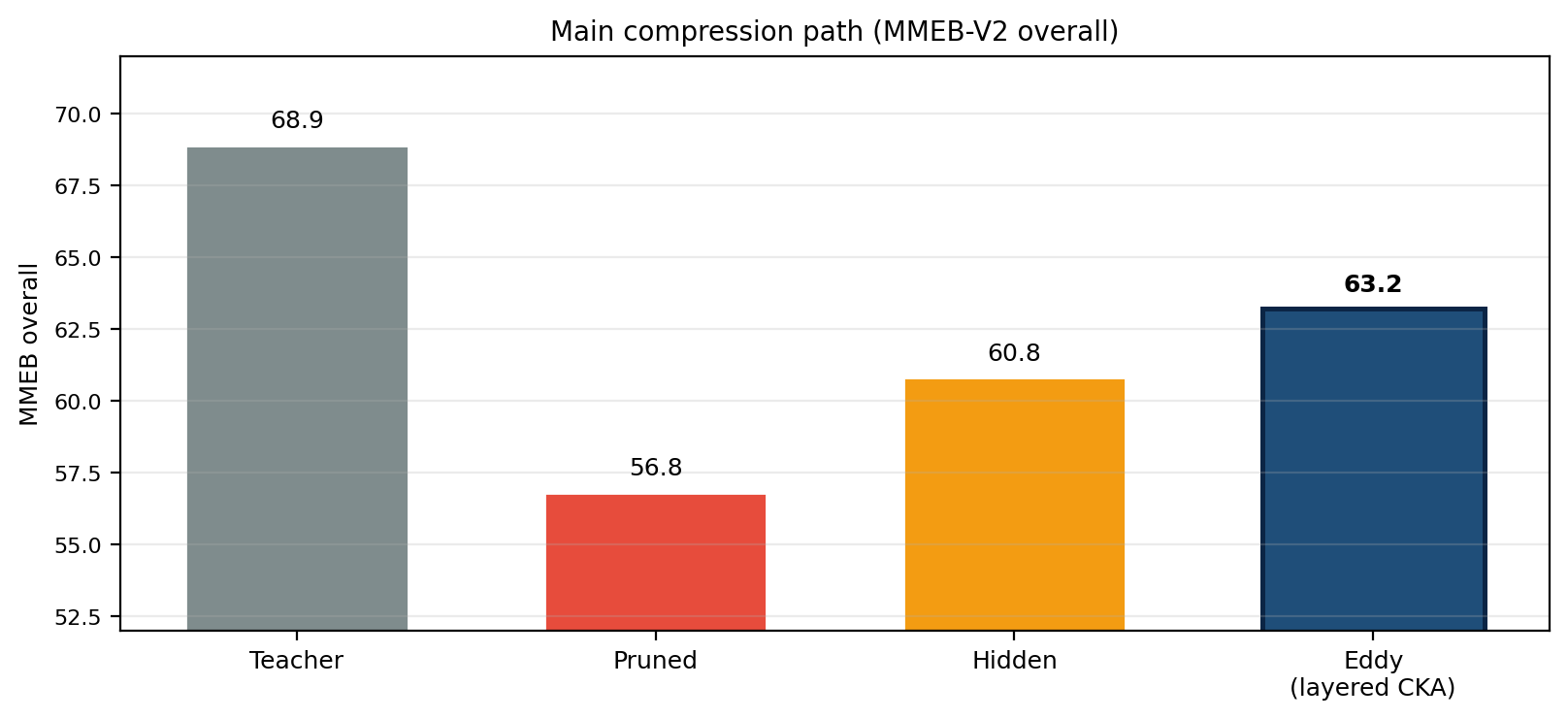}
  \caption{MMEB-V2 overall for the main compression path: teacher $\rightarrow$ pruned
           $\rightarrow$ hidden distill $\rightarrow$ Eddy (layered / attention-map CKA).}
  \label{fig:ablation}
\end{figure}

% Auto-generated ablation table
\begin{table}[H]
\centering
\small
\caption{MMEB-V2 distillation ablation (one row per training recipe). Overall = unweighted mean of per-task primary metrics (image/video: Hit@1; VisDoc: nDCG@5). $^\ddagger$\textbf{Hidden}: final hidden loss only---no mid-layer mapping losses. $^\star$\textbf{Layered / attention CKA}: Eddy-VL 1.9B release (mid losses on $\mathcal{M}$ + MRL final; best checkpoint). $^\dagger$\textbf{Layered} variants: same recipe as release but a different mid-layer objective (\S\ref{sec:training-objective}).}
\label{tab:ablation-pipeline}
\begin{tabular}{@{}lcccc@{}}
\toprule
Training recipe & Overall & Image & Video & VisDoc \\
\midrule
Qwen3-VL-Embedding-2B (teacher) & 68.9 & 71.0 & 55.2 & 76.1 \\
\textbf{Layered / attention CKA (Eddy release)}$^\star$ & 63.2 & 63.4 & 50.0 & 72.7 \\
Pruned baseline & 56.8 & 55.7 & 46.2 & 66.4 \\
Hidden distill (final hidden only)$^\ddagger$ & 60.8 & 60.8 & 48.6 & 69.9 \\
Layered / RCKA$^\dagger$ & 61.7 & 62.1 & 48.9 & 70.8 \\
Layered / STRUCTURE$^\dagger$ & 60.6 & 60.6 & 48.1 & 70.0 \\
Layered / mutual-kNN$^\dagger$ & 58.9 & 58.8 & 45.2 & 68.2 \\
\bottomrule
\end{tabular}
\end{table}

\subsection{Compositional retention}

\begin{figure}[H]
  \centering
  \includegraphics[width=0.88\textwidth]{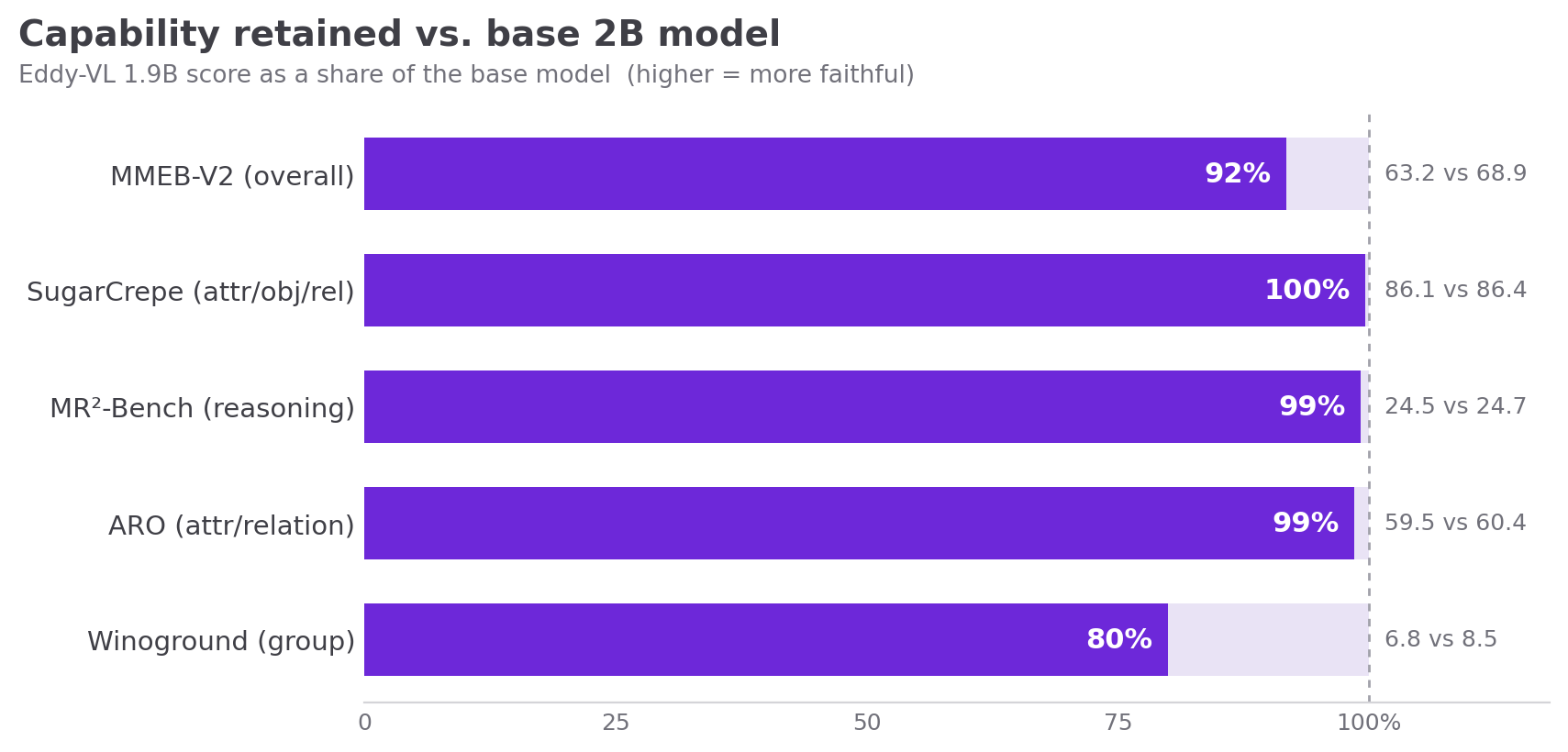}
  \caption{Capability retention vs.\ teacher on compositional benchmarks.}
  \label{fig:retention}
\end{figure}

SugarCrepe and MR\textsuperscript{2} recover to within $\sim$1 point of teacher;
Winoground group score is the lowest among the listed probes (6.8 vs.\ 8.5).

\paragraph{Interpretation.}
Fine-grained probes highlight how pruning and distillation affect compositional alignment
separately from MMEB-V2 task mixtures: Eddy recovers to near-teacher on SugarCrepe and
MR$^2$, while Winoground leaves the most headroom for improvement.

% ===========================================================================
\section{Conclusion}
\label{sec:conclusion}
% ===========================================================================

Eddy-VL 1.9B combines probe-driven depth pruning and layered CKA distillation to deliver a
1.93B multimodal embedder retaining $\sim$92\% of teacher MMEB score under our evaluation
protocol, with near-parity on compositional probes and $\sim$10\% lower forward latency.
Layered attention-map CKA outperformed hidden-only distillation and other mid-loss variants
in ablations on the combined Korean COCO + OurDataset mixture, yielding the release
score of 63.2.
Eddy-VL is an embedder for candidate surfacing---pair with a reranker for final ordering.

\paragraph{Limitations.}
\begin{itemize}[leftmargin=*]
  \item $\sim$8.3\% relative MMEB gap vs.\ teacher under our protocol ($-$5.7 absolute).
  \item Our MMEB re-evaluation $\neq$ Qwen leaderboard 73.2 (\S\ref{sec:eval-setup}).
  \item Training corpora are proprietary and not publicly released (AI Hub--licensed
        and in-house subsets; see \S\ref{sec:our-dataset}).
  \item Winoground remains the most challenging probe under our compression recipe.
\end{itemize}

\bibliographystyle{plain}
\bibliography{references}

\end{document}